\pdfoutput=1

\documentclass[11pt]{article}

\usepackage[]{ACL2023}

\usepackage{times}
\usepackage{latexsym}

\usepackage[T1]{fontenc}

\usepackage[utf8]{inputenc}

\usepackage{microtype}

\usepackage{booktabs, colortbl} 
\usepackage{multirow,makecell} 
\usepackage{multicol}
\usepackage{amsmath}
\usepackage{amssymb}
\usepackage{caption}
\usepackage{subcaption}
\usepackage{titlesec}
\usepackage{pgfplots}
\usepackage[normalem]{ulem} 
\usepackage{enumitem}
\usepackage{graphicx} 
\usepackage{hyperref}
\usepackage{svg}

\usepackage{inconsolata}

\title{Language-Independent Representations Improve Zero-Shot Summarization}

\author{Vladimir Solovyev\thanks{\phantom{--}Work done while at Karlsruhe Institute of Technology} \phantom{\and} Danni Liu \phantom{\and} Jan Niehues \\
        Karlsruhe Institute of Technology, Germany \\
        \texttt{vladimir.solovyev.90@gmail.com}, \texttt{\{danni.liu, jan.niehues\}@kit.edu}}

\begin{document}
\maketitle
\begin{abstract}
Finetuning pretrained models on downstream generation tasks often leads to catastrophic forgetting in zero-shot conditions.
In this work,
we focus on summarization and tackle the problem through the lens of language-independent representations.
After training on monolingual summarization,
we perform zero-shot transfer to new languages or language pairs.
We first show naively finetuned models are highly language-specific in both output behavior and internal representations, resulting in poor zero-shot performance.
Next, 
we propose query-key (QK) finetuning to decouple task-specific knowledge from the pretrained language generation abilities.
Then, 
after showing downsides of the standard adversarial language classifier,
we propose a balanced variant that more directly enforces language-agnostic representations.
Moreover, our qualitative analyses show removing source language identity correlates to zero-shot summarization performance.
Our code is openly available\footnote{\url{https://github.com/vladsolovyev/fairseq_summarization/tree/main/summarization_scripts}}.
\end{abstract}

\section{Introduction}

\begin{figure}
    \centering
    \includegraphics[trim={0.4cm 1cm 1.6cm 0.2cm},clip,width=\linewidth]{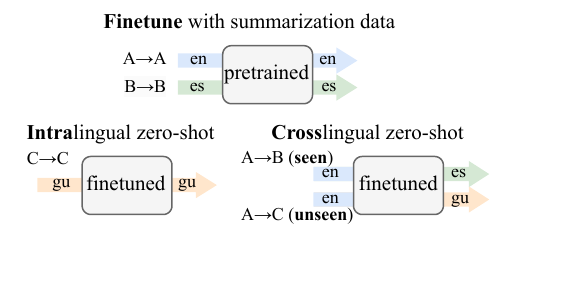}
    \caption{We finetune a pretrained model (e.g. mBART) on intralingual summarization data and test it on zero-shot intralingual and crosslingual summarization.}
    \label{fig:approach_overview}
\end{figure}

Pretrained multilingual models \cite{conneau-etal-2020-unsupervised,liu-etal-2020-multilingual-denoising,xue-etal-2021-mt5,lin-etal-2022-shot} have been established as promising sources of transfer learning,
where task-specific finetuning benefits from the general knowledge learned on diverse unsupervised data.
However,
due to data or computational constraints,
the task-specific data often only cover a limited subset of the languages in pretraining.
Therefore,
during finetuning
it is crucial to retain the knowledge of the pretrained model and to enable zero-shot transfer, i.e., performing the task on more languages covered by the pretrained model.
While zero-shot crosslingual transfer has shown very promising results on sequence classification or labeling problems \cite{pires-etal-2019-multilingual,DBLP:conf/nips/ConneauL19,wu-dredze-2019-beto},
it remains challenging for generation tasks \cite{ronnqvist-etal-2019-multilingual,vu-etal-2022-overcoming,li-murray-2023-zero} including summarization and translation.
A main obstacle is catastrophic forgetting \cite{DBLP:journals/neco/FrenchC02}, where languages supported by the pretrained model but not covered in the finetuning data are forgotten.
In this work, 
we use summarization as a testbed for various types of zero-shot generation.
As shown in \autoref{fig:approach_overview}, 
given a pretrained model
and intralingual summarization training data in some languages (A$\rightarrow$A, B$\rightarrow$B),
we aim for zero-shot \textit{intralingual} and \textit{crosslingual} summarization on new languages (C$\rightarrow$C) and language pairs (A$\rightarrow$B, A$\rightarrow$C) respectively.

To alleviate catastrophic forgetting, 
one line of work trains on additional unsupervised data \cite{maurya-etal-2021-zmbart,vu-etal-2022-overcoming,DBLP:journals/corr/abs-2311-09344}.
Besides the computational overhead,
this approach raises a theoretical question:
As the pretrained language model has already learned extensively on unsupervised data,
is it necessary to re-learn language modeling in task-specific finetuning?
We therefore explore a more challenging case of only using  
paired summarization data without relying on any unsupervised data.

We identify two challenges when generalizing summarization abilities to new languages.
First, 
decoupling the \textit{task}-specific knowledge from the \textit{language} generation abilities is essential.
In response,
we propose a new finetuning method based on query and keys, which is shown effective for both intralingual and crosslingual zero-shot setting.
For crosslingual zero-shot settings,
it is also crucial to decouple \textit{language} from \textit{content}, i.e., creating language-agnostic representations. 
This has been shown to facilitate zero-shot crosslingual generation in general \cite{pham-etal-2019-improving,wu-etal-2022-laft,DBLP:journals/corr/abs-2308-11466}.
There a prominent approach is adversarial training \cite{DBLP:journals/jmlr/GaninUAGLLML16,chen-etal-2018-adversarial},
where the model is trained to deceive a language classifier.
We show the existing formulation fails to fully achieve language-agnostic representations,
and improves it by explicitly incentivizing the model to deceive the classifier into a uniform class distribution.




\section{Current Models are Highly Language-Specific} \label{sec:current}

\begin{table}[t!]
\small
\centering
\setlength\tabcolsep{3pt} 
\begin{tabular}{cccccccccccc}
\toprule
\multicolumn{3}{c}{train on \textbf{en}} &&
\multicolumn{5}{c}{train on \text{en+es+ru}}
\\
\cmidrule{0-2}
\cmidrule{5-9}
\textbf{es-es} &
\textbf{ru-ru} &
\textbf{gu-gu} &&
\textbf{gu-gu} & 
\textbf{es-en} & 
\textbf{ru-en} &
\textbf{es-ru} & 
\textbf{tr-en} 
\\
0.2 &
2.3 &
13.4 &&
99.6 &
0.0 &
0.0 &
0.0 &
1.3 
\\
\bottomrule
\end{tabular}
\caption{\label{tab:lang_accuracy}
Proportion of generated summaries in the correct language (\%) under zero-shot conditions.
Codes: 
es (Spanish), ru (Russian),  gu (Gujarati), tr (Turkish).}
\end{table}
\begin{figure}
    \centering
    \includegraphics[trim={0.4cm 0.7cm 0.7cm 0.6cm},clip,width=0.7\linewidth]{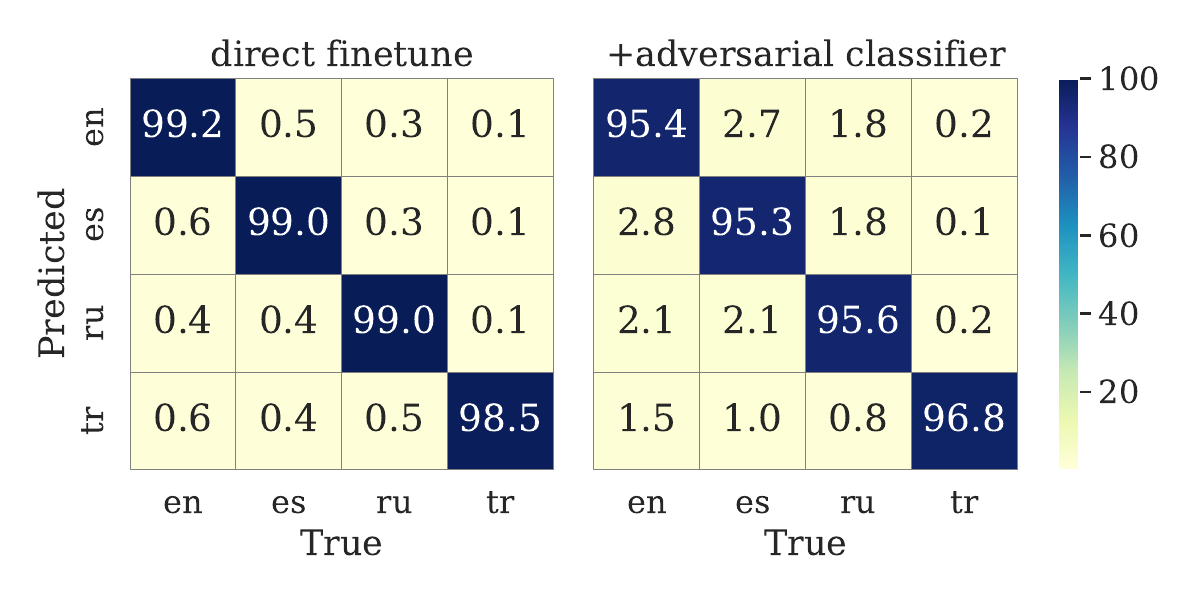}
    \caption{Accuracy of a probing classifier. Higher accuracy indicates more language-specific representations.}
\label{fig:probing_baseline_vs_vanilla_adv}
\end{figure}

We first show that 
naive finetuning makes the models highly language-specific in their \textit{output behavior} and \textit{internal representations}.
\autoref{tab:lang_accuracy} shows the proportion of outputs in the correct language after finetuning mBART on intralingual summarization data.
Finetuning on English only leads the model to forget its generation ability for other languages,
resulting in \textit{off-target generation} \cite{zhang-etal-2020-improving,pfeiffer-etal-2023-mmt5}, 
where a wrong target language, often one with supervised data, is generated.
Although multilingual training largely resolves off-target generation in intralingual settings\footnote{This is consistent with recent or concurrent findings \cite{DBLP:journals/corr/abs-2310-09917,pfeiffer-etal-2023-mmt5}.},
the problem persists for crosslingual generation.
As zero-shot crosslingual generation relies on language-agnostic representations, 
we test for this with a probing analysis \cite{DBLP:conf/iclr/AdiKBLG17}.
Specifically, 
we assess the difficulty of recovering the source language identity on the encoder output.
Given a trained model,
we train a token-level classifier for the input languages on the encoder outputs.\footnote{Details on the probing analysis are in \autoref{appx:probing_details}.}
As shown in \autoref{fig:probing_baseline_vs_vanilla_adv},
the classifier can almost perfectly recover the source language.
Even after explicitly encouraging language-agnostic representations with an adversarial language classifier \cite{DBLP:journals/corr/abs-1903-07091}, 
recovering the source language identity remains easy.

\section{Approaches}

\subsection{Decoupling Language from Task}
\paragraph{Query-Key (QK) Finetuning}
Prior works on zero-shot generation \cite{DBLP:conf/aaai/Chi0WWMH20,maurya-etal-2021-zmbart,li-etal-2021-multilingual} 
have highlighted the need for selective finetuning to mitigate forgetting,
where the consensus is updating the encoder and cross-attention weights only.
However, 
existing methods treat attention weights as a whole.
A closer look at the attention module reveals that,
only the value projections determine the basis of the upcoming transformations, 
whereas the query and key control how the inputs are aggregated.
We hypothesize that the value projections should be kept unchanged to prevent losing pretrained generation capabilities during finetuning.
In contrast, query and key are updated as adaptation to specific tasks.
Therefore, 
we propose a selective finetuning approach, 
which only updates the query and key projection weights of encoder self-attention and cross-attention.\footnote{
It is extendable to the parameter-efficient finetuning (PEFT) approach LoRA \cite{DBLP:conf/iclr/HuSWALWWC22} 
by placing the adapters on the query and key weights only.
In the experiments we do not compare to prominent PEFT approaches like prompt tuning \cite{lester-etal-2021-power} and LoRA, 
as prior works have shown they in their standard forms still suffer from catastrophic forgetting in finetuning \cite{vu-etal-2022-overcoming} or continual pretraining \cite{DBLP:journals/corr/abs-2401-03129}.}

\paragraph{Two-Step Finetuning}
For the more challenging case of crosslingual zero-shot summarization,
our approach is motivated by the fact that the task consists of two subtasks: translation and summarization.
We first finetune the pretrained model for translation\footnote{We do not use the finetuned mBART on translation \cite{DBLP:journals/corr/abs-2008-00401} as it can only translate from or into English.}.
Then we finetune again on intralingual summarization using our proposed query-key finetuning to retain its crosslingual capabilities acquired from translation in the first step.

\subsection{Decoupling Language from Content}
An adversarial language classifier is often used to decouple language from the semantic representations of input contents.
Most existing works use the cross-entropy loss \cite{DBLP:journals/corr/abs-1903-07091,mallinson-etal-2020-zero} and a gradient reversal layer \cite{DBLP:journals/jmlr/GaninUAGLLML16} 
to update the encoder weights in the opposite direction of the classifier accuracy.\footnote{More details in \autoref{appx:adv_details}}
A problem with the cross-entropy-based formulation is that it operates on \textit{single classes} and does not incentivize language-agnostic representations on the output \textit{distribution} level.
The adversarial classifier could potentially be shift all its predicted probability mass to another language,
achieving a low classification accuracy but leaving the representations still language-specific. 
Indeed as shown in \autoref{fig:probing_baseline_vs_vanilla_adv},
even after training with this objective,
a probing classifier can still easily learn to recover the source language identity.

\paragraph{Balanced Adversarial Language Classifier}
Given the drawback above, 
we propose a balanced adversarial objective.
Specifically, 
we train the encoder such that a language classifier is only able to predict an uniform distribution.
We achieve this by a modified adversarial loss based on the KL-divergence between the classifier output distribution and a uniform distribution: 
\begin{equation} \label{eq:loss2}
\mathcal{L}_\text{balanced\_adversarial} = D_{\text{KL}}(P_{\theta_{\text{classifier}}} \parallel U),
\end{equation}
where $P$ is the classifier output distribution on token level and $U=(\frac{1}{N}, \dots, \frac{1}{N})$ with $N$ being the number of languages to classify.

\paragraph{Residual Drop}
We further combine our approach  with residual drop \cite{liu-etal-2021-improving-zero}, 
a method proposed for machine translation that drops the residual connection of a middle encoder layer to reduce source language signals in the encoder output.

\section{Experiments and Results}
\vspace{-5pt}
\subsection{Experimental Setup}

\noindent
\textbf{Datasets}
We train on intralingual summarization data in English or \{English, Spanish, Russian\}.
We use XL-Sum \cite{hasan-etal-2021-xl} and WikiLingua \cite{ladhak-etal-2020-wikilingua} for experiments in \autoref{tab:partial_finetuning_results} and \autoref{tab:main_results} and respectively. 
The dataset details are in Appendix \ref{appx:dataset_stats}.
For the two-step finetuning, 
the translation data details are in Appendix \ref{appx:tranlsation_data}.

\paragraph{Data Conditions}
Besides the direct zero-shot condition, 
we compare to the following two data conditions:
\begin{itemize}[nolistsep,leftmargin=*]
    \item \textbf{Pipeline} approach translating into and from English:
    learn summarization on English only and 
    translate with NLLB-200 \cite{nllb}, 
    a recent open multilingual translation model. 
    Here we rely on English-only summarization as English has the most training data in both datasets, 
    which presumably yields the highest summarization quality.
    While this approach ensures that the outputs are in the right language, 
    the downsides are inference latency and translation error propagation.
    \item \textbf{Supervised}: 
    train on supervised data for the zero-shot directions as performance upper-bounds.
\end{itemize}

\paragraph{Baselines}
We compare our QK finetuning to:
\begin{itemize}[nolistsep,leftmargin=*]
    \item Encoder finetuning \cite{DBLP:conf/aaai/Chi0WWMH20}: It only updates the encoder weights to retain the pretrained generation capability, as the decoder is expected to be more responsible for generation.
    \item Layernorm and attention (LNA) finetuning \cite{li-etal-2021-multilingual}: It only finetunes: 1) layernorm, 2) encoder self-attention, and 3) cross-attention.
\end{itemize}

We also compare to the standard formulation of the adversarial language classifier \cite{DBLP:journals/corr/abs-1903-07091} based on the cross-entropy loss.

\paragraph{Training and Evaluation}
We initialize from the mBART \cite{liu-etal-2020-multilingual-denoising} model, which was pretrained on monolingual data of 25 languages.
Further training details are in \autoref{appx:training_details}.
To assess summarization quality, we use ROUGE \cite{lin-2004-rouge} and BERTScore \cite{DBLP:conf/iclr/ZhangKWWA20}.
We report ROUGE-L in the main text supply ROUGE-1/2 in \autoref{appx:detailed_exp_scores}.
We use BERTScore $F_{1}$ ($F_{\text{BERT}}$) following the authors' suggestions \cite{DBLP:conf/iclr/ZhangKWWA20}.
To measure the percentage of outputs in the correct languages, 
we use a language identifier \cite{lui-baldwin-2012-langid} and report accuracy (\%).

\begin{table*}[ht!]
\small
\centering
\setlength\tabcolsep{6pt} 
\begin{tabular}{llrrrrrrrrrrrrrrrrrrrrrrrr}
\toprule
\textbf{ID} &
\textbf{Model} &
\multicolumn{2}{c}{
\textbf{es}
} 
&& 
\multicolumn{2}{c}{\textbf{ru}}
&& 
\multicolumn{2}{c}{\textbf{gu}}
&& 
\multicolumn{2}{c}{\textbf{gu}}
\\
&&
\multicolumn{2}{c}{
(train on en)
} 
&& 
\multicolumn{2}{c}{
(train on en)}
&& 
\multicolumn{2}{c}{
(train on en)}
&& 
\multicolumn{2}{c}{
(train on en+es+ru)}
\\
\cmidrule{3-4} 
\cmidrule{6-7}
\cmidrule{9-10}
\cmidrule{12-13}
&& 
RG-L &
$F_{\text{BERT}}$ & 
& 
RG-L &
$F_{\text{BERT}}$ & 
& 
RG-L &
$F_{\text{BERT}}$ & 
& 
RG-L &
$F_{\text{BERT}}$ 
\\
\midrule
$(1)$ & 
Full ft.
& 5.4 & 66.0 & 
& 1.0 & 64.3 & 
& 1.2 & 59.1 & 
& 15.1 & 71.8 
\\
$(2)$ & 
Encoder ft. \cite{DBLP:conf/aaai/Chi0WWMH20}
& 18.4 & 70.8 & 
& 22.7 & 73.2 & 
& 14.5 & 71.7 & 
& 15.3 & 72.2 
\\
$(3)$ & 
``LNA'' ft. \cite{li-etal-2021-multilingual}
& 20.9 & 71.9 & 
& 21.6 & 72.7 & 
& 10.5 & 68.6 & 
& 16.0 & 72.6 
\\
$(4)$ & 
Query-key ft. (ours)
& \textbf{21.3} & \textbf{72.3} & 
& \textbf{23.4} & \textbf{73.6} & 
& \textbf{16.6} & \textbf{73.2} & 
& \textbf{16.5} & \textbf{73.1} 
\\
\midrule
$(5)$ &
Pipeline (translate to/from en)
& 20.7 & 72.1 & 
& 20.2 & 72.4 & 
& 13.6 & 72.1 & 
& 13.6 & 72.1 
\\
$(6)$ &
Supervised
& 25.0 & 74.0 & 
& 27.5 & 75.1 & 
& 19.3 & 74.2 & 
& 19.3 & 74.2 
\\
\bottomrule
\end{tabular}

\caption{\label{tab:partial_finetuning_results}
Zero-shot intralingual summarization results on XL-Sum.}
\end{table*}

\begin{table*}[ht!]
\small
\centering
\setlength\tabcolsep{0.25pt} 
\begin{tabular}{llrrrrrrrrrrrrrrrrrrrrrrrr}
\toprule
\textbf{ID} &
\textbf{Model} &
\multicolumn{2}{c}{\textbf{es-en}} 
&& 
\multicolumn{2}{c}{\textbf{ru-en}}
&& 
\multicolumn{2}{c}{\textbf{es-ru}}
&& 
\multicolumn{2}{c}{\textbf{avg. seen}}
&&
\multicolumn{2}{c}{\textbf{tr-en}}
&& 
\multicolumn{2}{c}{\textbf{en-tr}}
&& 
\multicolumn{2}{c}{\textbf{tr-tr}}
&& 
\multicolumn{2}{c}{\textbf{avg. unseen}}
\\
\cmidrule{3-4} 
\cmidrule{6-7}
\cmidrule{9-10} 
\cmidrule{12-13}
\cmidrule{15-16}
\cmidrule{18-19} 
\cmidrule{21-22} 
\cmidrule{24-25} 
&& 
RG-L &
$F_{\text{BERT}}$ & 
& 
RG-L &
$F_{\text{BERT}}$ &
& 
RG-L &
$F_{\text{BERT}}$ & 
& 
RG-L &
$F_{\text{BERT}}$ &
& 
RG-L &
$F_{\text{BERT}}$ &
& 
RG-L &
$F_{\text{BERT}}$ &
& 
RG-L &
$F_{\text{BERT}}$ &
& 
RG-L &
$F_{\text{BERT}}$ &
\\
\midrule
$(1)$
& Baseline zero-shot
& 2.2 & 67.8 &   
& 0.7 & 63.3 &   
& 0.6 & 64.6 &   
& \cellcolor[HTML]{d3d1d1} 1.2 
& \cellcolor[HTML]{d3d1d1} 65.2 &
& 4.6 & 62.9 &   
& 2.5 & 60.9 &   
& 18.0 & 71.5 &   
& \cellcolor[HTML]{d3d1d1} 8.4 
& \cellcolor[HTML]{d3d1d1}  65.1
\\
$(2)$
& Adv. classifier 
& 26.7 & 76.1 & 
& 25.3 & 75.7 & 
& 14.1 & 72.5 &
& \cellcolor[HTML]{d3d1d1} 22.0 
& \cellcolor[HTML]{d3d1d1} 74.8 &
& 26.1 & 75.2 & 
& 2.5 & 60.9 &
& 5.2 & 62.8 &
& \cellcolor[HTML]{d3d1d1} 11.3 
& \cellcolor[HTML]{d3d1d1} 66.3
\\
$(3)$
& Balanced adv. (ours)
& 27.2 & 76.4 &
& 25.6 & 75.8 & 
& 14.3 & 72.8 & 
& \cellcolor[HTML]{d3d1d1} 22.4 
& \cellcolor[HTML]{d3d1d1} 75.0 &
& 26.6 & 75.5 & 
& 2.6 & 60.9 &
& 3.2 & 61.1 &
& \cellcolor[HTML]{d3d1d1} 10.8 
& \cellcolor[HTML]{d3d1d1} 65.8
\\
$(4)$
& $(3) +$ residual drop
& 27.6 & \textbf{76.6} &
& \textbf{26.3} & \textbf{76.1} &
& \textbf{14.8} & 73.1 & 
& \cellcolor[HTML]{d3d1d1} \textbf{22.9} 
& \cellcolor[HTML]{d3d1d1} \textbf{75.3} &
& 25.7 & 75.2 &
& 2.5 & 61.0 &
& 2.3 & 60.8 &
& \cellcolor[HTML]{d3d1d1} 10.2 
& \cellcolor[HTML]{d3d1d1} 65.7
\\
$(5)$
& Two-step $+$ QK ft. (ours)
& \textbf{27.7} & 76.5 &
& \textbf{26.3} & \textbf{76.1} &
& \textbf{14.8} & \textbf{73.4} &
& \cellcolor[HTML]{d3d1d1} \textbf{22.9} 
& \cellcolor[HTML]{d3d1d1} \textbf{75.3} &
& \textbf{30.7} & \textbf{77.4} & 
& \textbf{16.7} & \textbf{71.3} & 
& \textbf{18.4} & \textbf{72.0} &
& \cellcolor[HTML]{d3d1d1} \textbf{21.9} 
& \cellcolor[HTML]{d3d1d1} \textbf{73.6}
\\
\midrule
$(6)$
& Pipeline
& 31.1 & 78.1 &
& 28.5 & 77.3 &
& 14.4 & 73.8 &
& \cellcolor[HTML]{d3d1d1} 24.7 
& \cellcolor[HTML]{d3d1d1} 76.4 &
& 34.1 & 78.7 &
& 18.7 & 73.1 &
& 18.5 & 73.2 &
& \cellcolor[HTML]{d3d1d1} 26.3 
& \cellcolor[HTML]{d3d1d1} 75.0
\\
$(7)$
& Supervised
& 31.4 & 78.1 &
& 29.4 & 77.5 &
& 18.0 & 75.2 &
& \cellcolor[HTML]{d3d1d1} 26.3 
& \cellcolor[HTML]{d3d1d1} 76.9 & 
& 34.5 & 78.8 &
& 20.7 & 73.2 &
& 26.2 & 75.4 &
& \cellcolor[HTML]{d3d1d1} 27.1 
& \cellcolor[HTML]{d3d1d1} 75.8
\\
\bottomrule
\end{tabular}
\vspace{-5pt}
\caption{\label{tab:main_results}
Zero-shot crosslingual summarization results on WikiLingua after training on \{en, es, ru\} intralingual data, 
grouped by \textit{seen} 
(new combinations of languages seen in finetuning) 
and \textit{unseen} 
(languages not in finetuning).}
\end{table*}
\subsection{Impact of Query-Key Finetuning}
The intralingual zero-shot results are in \autoref{tab:partial_finetuning_results} with detailed scores in \autoref{appx:detailed_exp_scores}.
Full finetuning (row $(1)$) on English-only data causes severe forgetting, 
where most of the output are in the wrong language,
which further harms summarization scores.

\paragraph{QK finetuning outperforms previous methods and pipeline approach:}
The proposed QK finetuning in row $(4)$ surpasses the two previous methods in row $(2)$ and $(3)$ by 0.4-2.1 ROUGE.
It is also the only approach consistently outperforming the translation-based pipeline in row $(5)$.
Moreover, 
the gap to the pipeline approach magnifies from high- to low-resource languages: For es, ru, gu, 
our QK finetuning leads by 2.9\%, 15.8\%, and 22.1\% ROUGE respectively.
This suggests that the two translation steps in the pipeline accumulates error that harms summarization quality,
especially on lower-resource languages.
Compared to the oracle condition with full supervised data (row $(6)$), 
the strongest zero-shot scores with our approach lies 2.7-4.1 ROUGE behind.
Given the difficulty of creating summarization data, 
this relatively small gap shows the potential of the zero-shot approach.

\paragraph{Comparison to multilingual training:}
Comparing the zero-shot results on Gujarati (gu), 
training multilingually on en+es+ru instead of English alone clearly prevents forgetting.
Even full finetuning in row $(1)$ almost always generates the correct target language.
Yet, QK finetuning still surpasses rows $(1)$-$(3)$.
Moreover,
its scores on gu when training on English only match those with multilingual training, 
suggesting it is a more data-efficient approach that does not rely on multilingual data.

\subsection{Impact of Removing Language Signals}
Despite its effectiveness so far, 
QK finetuning alone is not sufficient in crosslingual zero-shot settings.
The summarization scores\footnote{details in Appendix \ref{appx:detailed_exp_scores}} are very low in general as a result of off-target generation.
This leads to our next improvements on language-agnostic representations with results in \autoref{tab:main_results} with detailed scores in \autoref{appx:detailed_exp_scores}.

\begin{figure}
    \centering
    \includegraphics[trim={0.4cm 0.7cm 0.7cm 0.6cm},clip,width=0.7\linewidth]{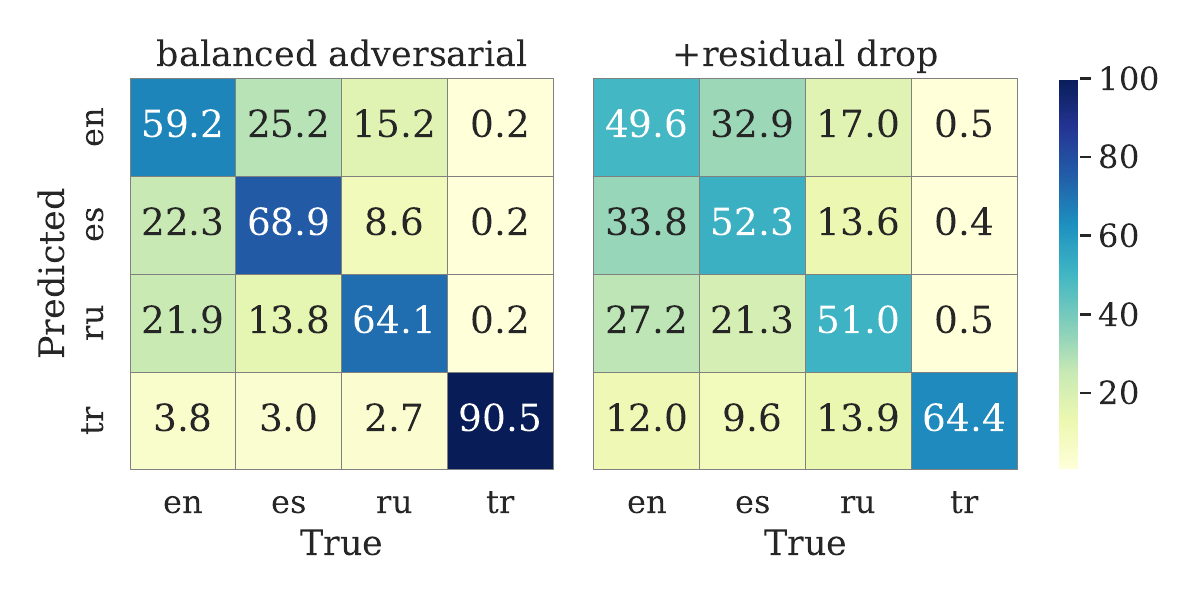}
    \caption{The models from rows $(3)$ and $(4)$ of \autoref{tab:main_results} have more language-agnostic representations and stronger zero-shot performance than those in \autoref{fig:probing_baseline_vs_vanilla_adv}.}
    \label{fig:probing_KLadv_residual}
\end{figure}

\paragraph{Removing language signals improves zero-shot performance for languages seen in finetuning:}
On language pairs where both the source and target are seen in finetuning (es-en, ru-en, es-ru),
our balanced adversarial classifier in row $(3)$ surpasses row $(2)$ by 0.4 ROUGE on average.
Combining it with the residual drop brings a further gain of 0.5 ROUGE.
However, 
these approaches are less effective on languages absent in the finetuning stage,
particularly on new target languages, 
as shown by the poor scores on en-tr and tr-tr.
Particularly, the intralingual summarization (tr-tr) quality degrades below the baseline.
This shows
that, to work on unseen languages, language-agnostic representations 
must be strengthened in target language generation.

\paragraph{Balanced adversarial classifier creates more language-agnostic representations:}
We probe the source language identity on the models trained with our balanced adversarial classifier and its combination with residual drop in \autoref{fig:probing_KLadv_residual}. 
Compare to \autoref{fig:probing_baseline_vs_vanilla_adv},
these two models' representations are clearly are language-agnostic.
The results show that language-agnostic representations are correlated to zero-shot cross-lingual summarization quality for languages seen in finetuning.

\subsection{Impact of Two-Step Finetuning}
Row $(5)$ of \autoref{tab:main_results} shows our two-step finetuning achieves strong zero-shot results for languages unseen in summarization finetuning.
As QK finetuning without the translation step was not capable of cross-lingual zero-shot generation,
we have evidence that the model retained knowledge from the crosslingual (translation) training.
Also, 
the two-step finetuning surpasses the pipeline approach on es-ru and tr-tr, where neither the source nor target is English, thereby needing translation twice.
This confirms the previous finding on translation error propagation harming summarization quality.



\vspace{-1.5pt}
\section{Conclusion}
In this work, 
we proposed two methods: 1) QK finetuning and 2) balanced adversarial language classifier to improve intralingual and crosslingual zero-shot summarization.
We presented evidence that language-independent representations facilitate zero-shot summarization,
in both intralingual and crosslingual forms.

We are curious to see the applicability of our methods to other generation tasks.
We are also curious about additional qualitative comparisons of language-specific and -independent representations.
In the current study, 
we only used probing analyses to assess language-specific versus language-independent representations. 
One way to supplement these analyses is to directly analyze the model hidden representations, e.g., compare the similarity between model hidden representations of different languages before and after applying the proposed approaches. 
This could for instance be achieved by Singular Vector Canonical Correlation (SVCCA) \cite{DBLP:conf/nips/RaghuGYS17}, 
which has been used to analyze multilingual representations for translation \cite{kudugunta-etal-2019-investigating,liu-etal-2021-improving-zero,sun-etal-2023-towards-deep}. 


\section*{Limitations}
This works has the following limitations:
\paragraph{Single Underlying Model}
All out experiments are based on mBART \cite{liu-etal-2020-multilingual-denoising}, specifically the variant pretrained on 25 languages.
Extending the current setup to mBART-50 which covers 50 languages can already provide wider language coverage for testing zero-shot inference.
Moreover, a further exploration with other pretrained models such as mT5 \cite{xue-etal-2021-mt5} or recent decoder-only large languages models \cite{bloom,llama2} could further validate the results.

\paragraph{Reliance on Translation Data}
Our two-step finetuning approach requires many-to-many translation data for the languages of interest. In extremely low-resource cases, we would need to create synthetic data by backtranslation \cite{sennrich-etal-2016-improving}, which requires more computational resources.

\paragraph{Lack of Multiple Experiment Runs}
Due to computational constraints, 
the scores in our experiments are reported from single experiment runs.
As a partial remedy, we use bootstrap resampling to derive confidence intervals of the reported scores and report the results in Appendix \ref{appx:detailed_exp_scores}.


\section*{Acknowledgement}
We thank the anonymous reviewers for helpful feedback.
Part of this work was performed on the HoreKa supercomputer funded by the
Ministry of Science, Research and the Arts Baden-Württemberg and by
the Federal Ministry of Education and Research.
Part of this work was supported by funding from the pilot program Core-Informatics of the Helmholtz Association (HGF).

\bibliography{anthology,custom}
\bibliographystyle{acl_natbib}

\appendix

\section{Details on Probing Analysis} \label{appx:probing_details}
The probing experiment aims to analyze the model hidden representations regarding the information they contain.
Here we are interested in the source language signals in the encoder outputs.
We freeze a trained model on the WikiLingua dataset \cite{ladhak-etal-2020-wikilingua}, 
and train a token-level classifier on the encoder outputs to recover the source language identity, 
where higher accuracy indicates more language-specific representations.
Specifically, 
we use a linear projection from the hidden dimension to the number of output classes, followed by a softmax activation.
For the output classes, 
we consider the four languages in the crosslingual experiments: English, Spanish, Russian, Turkish.
The classifier is trained on the same data as in the summarization task.

\section{Details on Adversarial Loss} \label{appx:adv_details}
With the standard cross-entropy-based adversarial classifier \cite{DBLP:journals/corr/abs-1903-07091,mallinson-etal-2020-zero}, 
the classifier itself is trained to optimize
\begin{equation} \label{eq:loss1}
\mathcal{L}_\text{classification} = -\sum_{c=1}^{L} y_c \text{log}(p_c),
\end{equation}
where $L$ is the number of languages, 
$y_c$ is a binary indicator based on whether the true language label is $c$, and $p_c$ is the predicted probability for language $c$.
To train the model to deceive the classifier, 
the adversarial loss is therefore the inverse of \autoref{eq:loss1}:
\begin{equation} \label{eq:loss2}
\mathcal{L}_\text{adversarial} = -\sum_{c=1}^{L} y_c \text{log}(1-p_c).
\end{equation}
However, 
the term will only be activated when $y_c$ is true, i.e., when the true label is $c$. 
This means when the classifier places all its probability mass on another language that is not $c$ (hence still highly language-specific), 
the adversarial loss will not update the model to change its representations.
This leaves the resulting language-specific representations unresolved.

\section{Dataset Statistics} 

\subsection{Dataset Splits} \label{appx:dataset_stats}

For the intralingual experiments, 
we train on XL-Sum \cite{hasan-etal-2021-xl}.
\autoref{tab:stats_xlsum} shows the dataset statistics.
\begin{table}[h]
\small
\centering
\setlength\tabcolsep{1pt} 
\begin{tabular}{llccccccccccc}
\toprule
&
\textbf{Split} &
\textbf{\# Samples} &
\textbf{Avg. input leng.} &
\textbf{Avg. output leng.}
\\
\midrule
English
& Train
& 302,627
& 459.9
& 22.3
\\
& Dev
& {\phantom-}{\phantom-}11,535
& 440.4
& 21.2
\\
& Test
& {\phantom-}{\phantom-}11,535
& 437.3
& 21.2
\\
Spanish
& Train
& {\phantom-}{\phantom-}35,633
& 723.5
& 29.4
\\
& Dev
& {\phantom-}{\phantom-}{\phantom-}4,763
& 766.5
& 27.4
\\
& Test
& {\phantom-}{\phantom-}{\phantom-}4,763
& 764.8
& 27.4
\\
Russian
& Train
& {\phantom-}{\phantom-}60,044
& 564.0
& 26.1
\\
& Dev
& {\phantom-}{\phantom-}{\phantom-}7,780
& 466.3
& 24.2
\\
& Test
& {\phantom-}{\phantom-}{\phantom-}7,780
& 465.3
& 24.2
\\
Gujarati
& Train
& {\phantom-}{\phantom-}{\phantom-}8,790
& 769.1
& 24.0
\\
& Dev
& {\phantom-}{\phantom-}{\phantom-}1,139
& 542.6
& 21.2
\\
& Test
& {\phantom-}{\phantom-}{\phantom-}1,139
& 529.9
& 21.7
\\
\bottomrule
\end{tabular}
\caption{\label{tab:stats_xlsum}
Dataset statistics on XL-Sum.
Training is done on English or \{English, Spanish, Russian\}.}
\end{table}
For the crosslingual experiments, 
we train on WikiLingua \cite{ladhak-etal-2020-wikilingua}.
\autoref{tab:stats_wikilingua} shows the dataset statistics.
\begin{table}[h]
\small
\centering
\setlength\tabcolsep{3pt} 
\begin{tabular}{llr|llr}
\toprule
\textbf{Lang.}&
\textbf{Split} &
\textbf{\# Samples} &
\textbf{Lang. pair} &
\textbf{Split} &
\textbf{\# Samples} 
\\
\midrule
\multicolumn{2}{l}{\textbf{Intralingual}} &&
\multicolumn{3}{l}{\textbf{Crosslingual}}
\\
en-en
& Train
& 95,517
& es-en
& Train
& 76,295
\\
& Dev
& {\phantom-}{\phantom-}3,000
& 
& Dev
& 3,000
\\
& Test
& 27,489
& 
& Test
& 21,726
\\
es-es
& Train
& 76,295
& ru-en
& Train
& 35,313
\\
& Dev
& {\phantom-}{\phantom-}3,000
& 
& Dev
& 3,000
\\
& Test
& 21,726
& 
& Test
& 9,962
\\
ru-ru
& Train
& 35,313
& es-ru
& Train
& 32,458
\\
& Dev
& 3,000
& 
& Dev
& 3,000
\\
& Test
& 7,780
& 
& Test
& 8,737
\\
tr-tr
& Train
& 8,790
& tr-en
& Train
& 3,052
\\
& Dev
& 1,139
& 
& Dev
& 438
\\
& Test
& 1,139
& 
& Test
& 874
\\
& 
& 
& en-tr
& Train
& 3,052
\\
& 
& 
& 
& Dev
& 438
\\
& 
& 
& 
& Test
& 874
\\
\bottomrule
\end{tabular}
\caption{\label{tab:stats_wikilingua}
Dataset statistics on WikiLingua.
Training is done on intralinuga data in English or \{English, Spanish, Russian\}.}
\end{table}
For both datasets, 
in training we exclude samples that have very short inputs (no more than 20 words or punctuation marks) or summaries (no more than 10 words or punctuation marks), as they likely have data quality issues.

\subsection{Details on Translation Data} \label{appx:tranlsation_data}
We use many-to-many data in all four languages evaluated in the crosslingual experiments:
English, Spanish, Russian, and Turkish.
To prepare the translation data, we parse the WikiLingua dataset by matching common intputs or outputs of different language pairs.
We iterate over samples in different language pairs and match samples that have the same input text or output summary in the
same language, 
but the corresponding output summary or input text is presented in different languages. 
By performing such matching, we generate translation data in
the same domain as used for summarization. 
A translation model trained with such data is capable of translating both short and long sequences.

\section{Training and Inference Details} \label{appx:training_details}

We implement our approaches in the \textsc{FairSeq} \cite{ott-etal-2019-fairseq} toolkit.

\begin{table*}[ht!]
\small
\centering
\setlength\tabcolsep{6pt} 
\begin{tabular}{llcccccccccccccccccccccc}
\toprule
\textbf{ID} &
\textbf{Model / Language} & 
RG-1 &
RG-2 &
RG-L &
$F_{\text{BERT}}$ & 
L-Acc.
\\
\midrule
& \textbf{es-es} (en-only train) \\
$(1)$ & 
Full ft. & 6.7/6.8/6.9 & 1.0/1.0/1.1 & 5.3/5.4/5.5 & 65.9/66.0/66.1 & 0.2 
\\
$(2)$ & 
Encoder ft. & 24.4/24.8/25.1 & 6.6/6.9/7.1 & 18.1/18.4/18.6 & 70.6/70.8/70.9 & 85.2 
\\
$(3)$ & 
``LNA'' ft. & 28.1/28.4/28.8 & 8.0/8.2/8.5 & 20.6/20.9/21.1 & 71.8/71.9/72.1 & 99.5 
\\
$(4)$ & 
Query-key ft.
& \textbf{28.3/28.6/29.0} & \textbf{8.6/8.8/9.1}
& \textbf{21.1/21.3/21.7} & \textbf{72.1/72.3/72.4} & 99.9 
\\
\midrule
$(5)$ &
Pipeline
& 27.8/28.1/28.5 & 8.0/8.2/8.5 & 20.4/20.7/21.0 & 72.0/72.1/72.3 & 100.0 
\\
$(6)$ &
Supervised
& 32.4/32.8/33.3 & 12.1/12.5/12.9 & 24.6/25.0/25.4 & 73.8/74.0/74.2 & 100.0 
\\
\\
& \textbf{ru-ru} (en-only train) \\
$(1)$ & 
Full ft. & 1.0/1.0/1.1 & 0.2/0.2/0.3 & 0.9/1.0/1.0 & 64.2/64.3/64.4 & 2.3 
\\
$(2)$ & 
Encoder ft. & 28.0/28.3/28.6 & 10.2/10.4/10.6 & 22.4/22.7/22.9 & 73.1/73.2/73.3 & 100.0 
\\
$(3)$ & 
``LNA'' ft. & 26.9/27.3/27.6 & 9.4/9.6/9.8 & 21.3/21.6/21.8 & 72.5/72.7/72.8 & 100.0 
\\
$(4)$ & 
Query-key ft. & 
\textbf{28.8/29.2/29.5} & \textbf{10.9/11.1/11.4} & \textbf{23.1/23.4/23.6} & \textbf{73.4/73.6/73.7} & 100.0 
\\
\midrule
$(5)$ &
Pipeline
& 25.0/25.3/25.6 & 7.9/8.1/8.3 & 20.0/20.2/20.5 & 72.3/72.4/72.5 & 100.0 
\\
$(6)$ &
Supervised
& 33.8/34.1/34.5 & 14.5/14.7/15.0 & 27.2/27.5/27.8 & 74.9/75.1/75.2 & 100.0 
\\
\\
& \textbf{gu-gu} (en-only train)  \\
$(1)$ & 
Full ft. & 1.2/1.3/1.5 & 0.2/0.3/0.3 & 1.1/1.2/1.4 & 59.0/59.1/59.3 & 13.4 
\\
$(2)$ & 
Encoder ft. & 15.1/15.8/16.6 & 4.3/4.8/5.3 & 13.8/14.5/15.2 & 71.4/71.7/72.0 & 100.0 
\\
$(3)$ & 
``LNA'' ft. & 11.2/11.7/12.4 & 2.8/3.2/3.6 & 9.9/10.5/11.1 & 68.3/68.6/68.9 & 99.8 
\\
$(4)$ & 
Query-key ft. &
\textbf{17.5/18.3/19.1}& \textbf{5.2/5.8/6.4} & \textbf{15.9/16.6/17.3} & \textbf{72.9/73.2/73.6} & 100.0 
\\
\midrule
$(5)$ &
Pipeline & 14.6/15.2/15.7 & 2.8/3.2/3.5 & 13.0/13.6/14.1 & 71.9/72.1/72.4 & 100.0
\\
$(6)$ &
Supervised & 20.8/21.5/22.4 & 7.1/7.7/8.4 & 18.5/19.3/20.1 & 73.9/74.2/74.5 & 100.0
\\
\\
& \textbf{gu-gu} (multi. train) \\
$(1)$ & 
Full ft. & 16.1/16.9/17.6 & 4.3/4.9/5.4 & 14.3/15.1/15.8 & 71.5/71.8/72.0 & 99.6
\\
$(2)$ & 
Encoder ft. & 16.0/16.8/17.5 & 4.4/4.9/5.4 & 14.6/15.3/15.9 & 71.9/72.2/72.5 & 100.0
\\
$(3)$ & 
``LNA'' ft. & 17.1/17.8/18.5 & 5.1/5.6/6.1 & 15.3/16.0/16.8 & 72.3/72.6/72.8 & 100.0
\\
$(4)$ & 
Query-key ft. &
\textbf{17.4/18.2/19.0} & \textbf{5.6/6.2/6.8} & \textbf{15.7/16.5/17.3} & \textbf{72.8/73.1/73.4} & 100.0
\\
\midrule
$(5)$ &
Pipeline & 14.6/15.2/15.7 & 2.8/3.2/3.5 & 13.0/13.6/14.1 & 71.9/72.1/72.4 & 100.0
\\
$(6)$ &
Supervised & 20.8/21.5/22.4 & 7.1/7.7/8.4 & 18.5/19.3/20.1 & 73.9/74.2/74.5 & 100.0
\\
\bottomrule
\end{tabular}

\caption{\label{tab:intralingual_full}
Full zero-shot intralingual summarization results on XL-Sum calculated using 95\% bootstrap confidence intervals (results are presented as 0.025/0.5/0.975 percentiles).}
\end{table*}
\begin{table}[t]
\small
\centering
\setlength\tabcolsep{7.5pt} 
\begin{tabular}{llrrrrrrrrrrrrrrrrrrrrrrrr}
\toprule
& 
es-en &
ru-en &
es-ru &
tr-en & 
en-tr \\
\midrule
ROUGE-L
&
2.2 &
0.7 &
0.5 &
2.3 &
2.5 
\\
L-Acc. &
0.0 &
0.0 &
0.0 &
0.0 &
0.0 \\
\bottomrule
\end{tabular}
\caption{\label{tab:qk_alone}
Results of QK finetuning alone (without two-step finetuning) under the crosslingual zero-shot setup.}
\end{table}
\begin{table*}[ht!]
\small
\centering
\setlength\tabcolsep{6pt} 
\begin{tabular}{llcccccccccccccccccccccc}
\toprule
\textbf{ID} &
\textbf{Model / Language} & 
RG-1 &
RG-2 &
RG-L &
$F_{\text{BERT}}$ & 
L-Acc.
\\
\midrule
& \textbf{es-en}\\
$(1)$ & 
Baseline zero-shot
& 2.3/2.4/2.4 & 0.1/0.1/0.1 & 2.1/2.2/2.2 & 67.8/67.8/67.9 & 0.0
\\
$(2)$ & 
Adv. classifier & 33.2/33.4/33.6 & 10.4/10.5/10.6 & 26.5/26.7/26.8 & 76.1/76.1/76.2 & 98.5
\\
$(3)$ & 
Balanced adv. classifier & 33.9/34.1/34.3 & 10.8/11.0/11.1 & 27.1/27.2/27.4 & 76.3/76.4/76.4 & 99.4
\\
$(4)$ & 
$(3)+$ residual drop
& \textbf{34.6/34.8/34.9} & 11.2/11.3/11.5
& 27.5/27.6/27.8 & \textbf{76.5/76.6/76.6} & 99.7
\\
$(5)$ &
Two-step $+$ QK ft. 
& 34.2/34.4/34.5 & \textbf{11.3/11.5/11.7}
& \textbf{27.5/27.7/27.9} & 76.4/76.5/76.6 & 98.4
\\
\midrule
$(6)$ &
Pipeline & 37.7/37.9/38.1 & 14.3/14.5/14.6 & 31.0/31.1/31.3 & 78.1/78.1/78.2 & 99.7
\\
$(7)$ &
Supervised & 38.2/38.4/38.6 & 14.6/14.8/14.9 & 31.2/31.4/31.6 & 78.1/78.1/78.2 & 99.8
\\
\\

& \textbf{ru-en}\\
$(1)$ & 
Baseline zero-shot
& 0.7/0.7/0.7 & 0.1/0.1/0.1 & 0.6/0.7/0.7 & 63.3/63.3/63.3 & 0.0
\\
$(2)$ & 
Adv. classifier & 31.6/31.9/32.2 & 9.7/9.9/10.1 & 25.1/25.3/25.5 & 75.6/75.7/75.7 & 99.6
\\
$(3)$ & 
Balanced adv. classifier & 31.9/32.2/32.5 & 10.0/10.2/10.4 & 25.4/25.6/25.9 & 75.7/75.8/75.9 & 99.8
\\
$(4)$ & 
$(3)+$ residual drop
& \textbf{32.9/33.1/33.4} & 10.4/10.6/10.8 & \textbf{26.0/26.3/26.5} & \textbf{76.0/76.1/76.2} & 99.9
\\
$(5)$ &
Two-step $+$ QK ft. 
& 32.6/32.8/32.9 & \textbf{10.6/10.7/10.8} & \textbf{26.1/26.3/26.4} & \textbf{76.0/76.1/76.3} & 99.6
\\
\midrule
$(6)$ &
Pipeline
& 34.4/34.6/34.9 & 12.0/12.2/12.4 & 28.2/28.5/28.7 & 77.2/77.3/77.4 & 99.7
\\
$(7)$ &
Supervised
& 35.7/36.0/36.3 & 12.9/13.2/13.4 & 29.2/29.4/29.7 & 77.4/77.5/77.5 & 99.7
\\
\\

& \textbf{es-ru}\\
$(1)$ & 
Baseline zero-shot
& 0.5/0.6/0.6 & 0.1/0.1/0.1 & 0.5/0.6/0.6 & 64.6/64.6/64.7 & 0.0
\\
$(2)$ & 
Adv. classifier & 16.6/16.9/17.1 & 3.9/4.1/4.2 & 13.8/14.1/14.3 & 72.4/72.5/72.6 & 97.6
\\
$(3)$ & 
Balanced adv. classifier & 17.1/17.3/17.5 & 4.1/4.3/4.4 & 14.1/14.3/14.5 & 72.7/72.8/72.9 & 99.9
\\
$(4)$ & 
$(3)+$ residual drop
& 17.6/17.8/18.0 & 4.3/4.5/4.6 & \textbf{14.5/14.8/15.0} & 73.0/73.1/73.2 & 100.0
\\
$(5)$ &
Two-step $+$ QK ft. 
& \textbf{17.4/17.6/17.8} & \textbf{4.5/4.6/4.6} & \textbf{14.7/14.8/14.9} & \textbf{73.2/73.4/73.6} & 98.4
\\
\midrule
$(6)$ &
Pipeline
& 16.4/16.7/16.9 & 3.5/3.7/3.8 & 14.2/14.4/14.6 & 73.7/73.8/73.8 & 100.0
\\
$(7)$ &
Supervised
& 20.8/21.0/21.3 & 6.0/6.1/6.3 & 17.7/18.0/18.2 & 75.1/75.2/75.3 & 100.0
\\
\\

& \textbf{tr-en}\\
$(1)$ & 
Baseline zero-shot
& 4.4/5.0/5.6 & 0.9/1.1/1.4 & 4.1/4.6/5.1 & 62.6/62.9/63.1 & 1.6
\\
$(2)$ & 
Adv. classifier & 31.0/32.0/33.0 & 10.0/10.7/11.4 & 25.3/26.1/26.9 & 74.9/75.2/75.5 & 98.9
\\
$(3)$ & 
Balanced adv. classifier &
31.7/32.6/33.6 & 10.3/11.0/11.7 & 25.7/26.6/27.4 & 75.2/75.5/75.8 & 99.8
\\
$(4)$ & 
$(3)+$ residual drop & 31.2/32.1/33.0 & 9.7/10.3/11.0 & 24.9/25.7/26.5 & 74.9/75.2/75.5 & 99.1
\\
$(5)$ &
Two-step $+$ QK ft. 
& \textbf{38.3/38.6/38.8} & \textbf{14.4/14.6/14.7} & \textbf{30.5/30.7/30.9} & \textbf{77.1/77.4/77.6} & 99.7
\\
\midrule
$(6)$ &
Pipeline
& 39.9/40.9/41.8 & 16.1/17.0/17.8 & 33.2/34.1/35.0 & 78.4/78.7/79.0 & 99.4
\\
$(7)$ &
Supervised
& 40.4/41.4/42.5 & 17.1/18.1/19.1 & 33.5/34.5/35.5 & 78.5/78.8/79.1 & 99.4
\\
\\

& \textbf{en-tr}\\
$(1)$ & 
Baseline zero-shot
& 2.4/2.7/3.0 & 0.4/0.5/0.5 & 2.3/2.5/2.7 & 60.8/60.9/61.1 & 0.0
\\
$(2)$ & 
Adv. classifier & 2.5/2.8/3.0 & 0.4/0.5/0.6 & 2.3/2.5/2.8 & 60.7/60.9/61.0 & 0.0
\\
$(3)$ & 
Balanced adv. classifier & 2.5/2.8/3.1 & 0.4/0.5/0.6 & 2.4/2.6/2.8 & 60.8/60.9/61.1 & 0.0
\\
$(4)$ & 
$(3)+$ residual drop & 2.4/2.7/3.0 & 0.4/0.5/0.6 & 2.3/2.5/2.7 & 60.8/61.0/61.1 & 0.0
\\
$(5)$ &
Two-step $+$ QK ft. 
& \textbf{20.2/20.4/20.6} & \textbf{5.7/5.8/5.8} & \textbf{16.5/16.7/16.9} & \textbf{71.0/71.3/71.5} & 98.7
\\
\midrule
$(6)$ &
Pipeline
& 20.1/20.9/21.6 & 5.1/5.5/6.0 & 18.0/18.7/19.4 & 72.8/73.1/73.4 & 99.8
\\
$(7)$ &
Supervised
& 22.7/23.7/24.8 & 7.4/8.0/8.7 & 19.7/20.7/21.5 & 72.9/73.2/73.5 & 100.0
\\
\\

& \textbf{tr-tr}\\
$(1)$ & 
Baseline zero-shot & 20.0/20.9/21.7 & 5.5/6.0/6.5 & 17.4/18.0/18.8 & 71.2/71.5/71.8 & 96.4
\\
$(2)$ & 
Adv. classifier & 5.3/5.7/6.2 & 1.0/1.1/1.3 & 4.7/5.2/5.6 & 62.6/62.8/63.0 & 10.6
\\
$(3)$ & 
Balanced adv. classifier & 3.2/3.5/3.8 & 0.5/0.5/0.7 & 3.0/3.2/3.5 & 60.9/61.1/61.3 & 0.1
\\
$(4)$ & 
$(3)+$ residual drop & 2.2/2.5/2.7 & 0.3/0.4/0.5 & 2.1/2.3/2.5 & 60.6/60.8/60.9 & 0.0
\\
$(5)$ &
Two-step $+$ QK ft. 
& \textbf{22.9/23.1/23.3} & \textbf{6.9/7.0/7.1} & \textbf{18.2/18.4/18.6} & \textbf{71.7/72.0/72.2} & 100.0
\\
\midrule
$(6)$ &
Pipeline & 19.9/20.7/21.5 & 5.4/5.8/6.3 & 17.7/18.5/19.2 & 73.0/73.2/73.5 & 99.8
\\
$(7)$ &
Supervised
& 29.2/30.3/31.3 & 11.4/12.3/13.0 & 25.3/26.2/27.2 & 75.2/75.4/75.7 & 99.7
\\

\bottomrule
\end{tabular}

\caption{\label{tab:crosslingual_full}
Full zero-shot crosslingual summarization results on WikiLingua calculated using 95\% bootstrap confidence
intervals (results are presented as 0.025/0.5/0.975 percentiles).}
\end{table*}

\paragraph{Training}
We initialized from the pretrained mBART model\footnote{We use the 610M \texttt{mbart.CC25} model from \url{https://github.com/facebookresearch/fairseq/blob/main/examples/mbart/README.md\#pre-trained-models}.} \cite{liu-etal-2020-multilingual-denoising}.
The word embeddings are frozen due to initial favourable results in zero-shot settings.
We use the Adam optimizer \cite{adam} with betas (0.9, 0.999) and eps 1e-8. 
We use 
weight decay of 0.01, 
start learning rate of 2e-5 
and end end learning rate of 5e-9.
Dropout is set to 0.1. 
We use the development set of the same languages as in training for early stopping.
All models are trained on an Nvidia Titan RTX GPU with 24GB memory.


\paragraph{Inference}
When decoding, we use a beam size of 5.
The length penalty is 0.6 and 1.0 for intralingual and crosslingual experiments respectively.
For the translation model in the pipeline approach, 
we use the distilled NLLB-200 model \cite{nllb} with 600M parameters.

\paragraph{Evaluation}
On XL-Sum, we follow the original dataset paper \cite{hasan-etal-2021-xl} and use the Multilingual ROUGE Scoring from \href{https://github.com/csebuetnlp/xl-sum/tree/master/multilingual_rouge_scoring}{{here}}.

\section{Detailed Experiment Scores} \label{appx:detailed_exp_scores}

\paragraph{Detailed Intralingual Results}
The detailed results for \autoref{tab:partial_finetuning_results} with ROUGE-1 and ROUGE-2 are in \autoref{tab:intralingual_full} with RG standing for ROUGE.

\paragraph{Detailed Crosslingual Results}
The detailed results for \autoref{tab:main_results} are in \autoref{tab:crosslingual_full}.

\paragraph{QK Finetuning in Crosslingual Settings}
QK finetuning alone is not sufficient in crosslingual zero-shot settings.
The scores are in \autoref{tab:qk_alone}.

\end{document}